%% file: main.tex
          [2001/04/25 1.1 (PWD)]
\documentclass[a4paper,twocolumn]{esapub2005} 
\usepackage{times}
\usepackage[numbers]{natbib}
\usepackage{url}
\usepackage{siunitx}
\usepackage{bbm}
\usepackage{amsfonts} 
\usepackage{pdfpages}
\usepackage{import}
\usepackage{lmodern}
\usepackage{comment}
\usepackage[super]{nth}
\usepackage{booktabs}
\newcommand{\ra}[1]{\renewcommand{\arraystretch}{#1}}

\title{Linear Model Predictive Control for a planar free-floating platform: A comparison of binary input constraint formulations}
\author[1,2]{Franek Stark}
\author[2,4]{Shubham Vyas}
\author[3]{Georg Schildbach}
\author[2,4]{Frank Kirchner}
\affil[1]{Robotics and Autonomous Systems, Universität zu Lübeck, Germany, franek.stark@student.uni-luebeck.de}
\affil[2]{Robotics Innovation Centre (RIC), DFKI GmbH, Germany, \small{\{franek.stark, shubham.vyas, frank.kirchner\}@dfki.de}}
\affil[3]{\normalsize Institute for Electrical Engineering in Medicine, \small Universität zu Lübeck, Germany, georg.schildbach@uni-luebeck.de}
\affil[4]{\normalsize AG Robotik, University of Bremen, Germany}


\usepackage{graphicx}
\usepackage{amsmath}
\usepackage[disable]{todonotes}

\input{acronyms}

\usepackage{layouts}
\begin{document}

\maketitle

\begin{abstract}
This work develops a first \acrlong{mpc} for \acrlong{esa}'s 3-dof free-floating platform. The challenges of the platform are the on/off thrusters, which cannot be actuated continuously and which are subject to certain timing constraints. 
This work compares penalty-term, \acrlong{lcc}, and classical \acrlong{mi} formulations in order to develop a controller that natively handles binary inputs. Furthermore, linear constraints are proposed which enforce the timing constraints.
Only the \acrlong{mi} formulation turns out to work sufficiently. Hence, this work develops a new \acrlong{mimpc} on the decoupled model of the platform. Feasibility analysis and simulation results show that for a short enough prediction horizon, this controller can (sub)optimally stabilize and control the system under consideration of the constraints in real-time. 
\end{abstract}

\section{Introduction}
Satellite control to do rendezvous maneuvers, e.g., in the field of space debris removal, recently gained attention \cite{vyas_post-capture_2022}. \todo[inline]{source!}
To test such controllers and subsystems, the \acrfull{esa}'s \acrfull{orgl} features a $9\times5$~m flat floor \citep{zwick_orgl_2018}. On top, the \SI{220}{kg} heavy air-bearing platform \acrshort{reacsa} simulates a satellite \citep{bredenbeck_trajectory_2022}.
\acrshort{reacsa} is equipped with a cold gas propulsion system and a reaction wheel. The latter exerts a precise desired torque to the system. The eight thrusters are mounted in a way to be able to control linear and angular acceleration.
Like a real satellite or system that uses rocket motors or reaction wheels, this experimental platform is subject to certain input constraints. In \acrshort{reacsa}'s case, these constraints are rather strict: Besides the reaction wheel's minimum and maximum angular velocity, the thrusters can only be on/off actuated. Thus, a thruster can either deliver its full thrust or no thrust. In addition, the thrusters are subject to certain time restrictions: Once a thruster has been switched on, it must remain switched on for a minimum time $t_{\text{on,min}}$ to exert repeatable and reliable thrust. Also, it must not exceed a maximum activation time $t_{\text{on,max}}$ due to the requirements imposed by the pneumatic system. After an activation phase, a thruster must have an off period $t_{\text{off,min}}$, while the buffer is filled up.\\
The current controller is a \acrfull{tvlqr} with preliminary trajectory optimization. However, it does not know about the system constraints and hence, in reality, performs poorly \citep{bredenbeck_trajectory_2022}. By enabling optimal control while respecting any constraint, developing an \acrfull{mpc} is hence the next step.

Existing works on \acrshort{mpc} for satellites with binary thrusters can be divided into different classes: 
First, continuous \acrshort{mpc} formulations treat the control input as continuous and convert it into binary values, e.g., by using a Delta-Sigma modulator \citep{virgili-llop_experimental_2018}. 
Or, like the \acrshort{mpc} presented by \citet{arantes_optimal_2009}, output continuous \acrshort{pwm} parameters, which assumes one firing cycle per prediction step and cannot flexibly allocate binary thruster values.\\
Since optimal control by exploiting perfect thruster allocation under consideration of the timing constraints is desired, this work focuses on the second class: formulations that have direct binary outputs.

Enforcing an input to be a binary variable makes the controller's underlying optimization problem significantly complex, since a non-convexity is introduced. The current state-of-the-art literature tackles this optimization problem using three different methods:
\begin{enumerate}
    \item A widely used approach is to define respective variables as integer variables and formulate the problem as a \acrfull{mip}.
    \item The binary variables are assumed to be continuous, and the problem is a \acrshort{qp}. To prioritize binary values, an additional quadratic penalty term is used \citep{abudia_switched_2020}.
    \item Good results have been achieved with so called \acrfull{mpcc} in the area of contact-implicit trajectory optimization \citep{posa_direct_2014}. Here, a quadratic non-convex constraint, called \acrshort{lcc}, enforces binary values.
\end{enumerate}
While for the \acrshort{mip} an appropriate solver is necessary, the \acrshort{qp} can be solved by any (non-convex) \acrshort{qp} solver. A \acrshort{mpcc} can be solved with any non-linear solver. However, problems with binary variables are NP-hard, and solvability in real-time depends strongly on the problem \citep{leeuwen_algorithms_1990}. \todo[inline]{true? otherwise, reformulate maybe to }

For the \acrshort{milp} approach, there are already works in the space and satellite context, but these are mostly concerned with attitude control and do not have such strict timing constraints on their inputs \citep{vieira_attitude_2011, doman_control_2007}.
And even of these, only a very few have implemented a real-time capable controller, like \citep{leomanni_mpc-based_2013, sopasakis_hybrid_2015}. An \acrshort{mpc} for a very similar platform without timing constraints is developed in \citet{khayour_active_2020}. Compared to classical controllers, it performs better, but is not real-time capable.
For the \acrshort{lcp} and quadratic-cost-based formulation, to the best of the author's knowledge, there are no results yet for a similar application.
\todo[inline]{add more references, especially the one which has the similar (cable) problem.}

The goal of this work is to find a \acrshort{mpc} formulation that knows about both the binary and the timing constraints and can solve the optimization problem within a reasonable time. Hence, the three different binary constraint formulations are compared on a simplified model of \acrshort{reacsa}. An \acrshort{mpc} for this is developed and tested in simulation.

This paper is organized as follows: Section~\ref{sec:system_desc} describes the system model used and gives some theoretical considerations about the system's limit cycle. Section~\ref{sec:mpc} introduces the \acrshort{mpc} formulations. The results are described in section~\ref{sec:results} followed by the summary and conclusion in section~\ref{sec:summary_conc}.
\section{System description}\label{sec:system_desc}
In this section, the used model of \acrshort{reacsa} is presented and analyzed for its limit cycle behavior.
\subsection{Dynamic model}
\begin{figure}[htpb]
    \centering
     \includegraphics[width=0.7\linewidth]{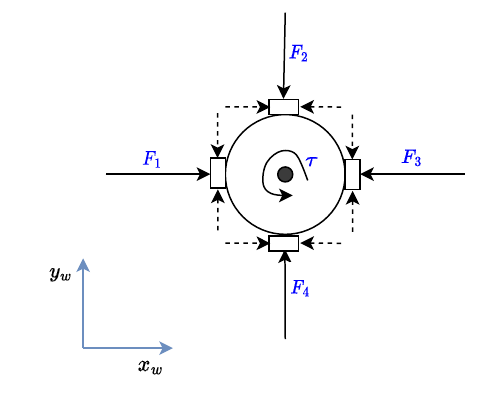}
    \caption{Schematics of the modeled system. Pairs of thruster forces (dashed) that apply thrust in the same direction are lumped together to one single (virtual) thruster which applies force $F_i$. Rotation is applied by the torque source~$\tau$.}
    \label{fig:system}
\end{figure}

Fig.~\ref{fig:system} shows the model of \acrshort{reacsa} used in this work. The two respective thrusters that accelerate in the same direction are combined into one force $F_i$. 
Therefore, the binary number of binary inputs is reduced to four, and binary inputs don't introduce a torque. 
Orientation changes are applied by the torque source $\tau$. Hence, this model is referred to as the decoupled model. Note that in this work the reaction wheel with its limits is not modeled.

The state vector $\mathbf{x}$ contains the linear and angular position and velocity in world coordinates:
\begin{equation}
    \mathbf{x} = \begin{bmatrix}
        x&y&\theta&\dot{x}&\dot{y}&\dot{\theta}
    \end{bmatrix}^T
\end{equation}
The system input vector $\mathbf{u}$ contains the applied torque $\tau$ and the four binary inputs $u_{1,...,4} \in \left\{0,1\right\}$:
\begin{equation}
        \mathbf{u} = \begin{bmatrix}
        \tau&u_1&u_2&u_3&u_4
    \end{bmatrix}^T
\end{equation}
The system dynamics are given by:
\begin{gather}
    \dot{\mathbf{x}} = f(\textbf{x}, \textbf{u})=\underbrace{\begin{bmatrix}
    \multicolumn{3}{c}{\mathbf{0}^{3\times3}}&\multicolumn{3}{c}{\mathbf{I}^{3\times3}}\\
    \multicolumn{6}{c}{\mathbf{0}^{3\times6}}\\
    \end{bmatrix}}_\mathbf{A} \mathbf{x} \nonumber\\+ \underbrace{\begin{bmatrix}
         \multicolumn{5}{c}{\mathbf{0}^{3\times5}}\\
        0&c\theta\frac{2F_\text{n}}{m}&c\theta\frac{-2F_\text{n}}{m}&s\theta\frac{2F_\text{n}}{m}&s\theta\frac{-2F_\text{n}}{m}\\[2pt]
        0&s\theta\frac{2F_\text{n}}{m}&s\theta\frac{-2F_\text{n}}{m}&c\theta\frac{-2F_\text{n}}{m}&c\theta\frac{2F_\text{n}}{m}\\
                 \frac{1}{I_{zz}}& \multicolumn{4}{c}{\mathbf{0}^{1\times4}}\\
    \end{bmatrix}}_{\mathbf{B}(\theta)}\mathbf{u}\label{eq:sys_dyn}
\end{gather}
Where $s\theta, c\theta$ denote the sinus and cosine of the system's orientation, $F_\text{n}$ denotes the force applied by a thruster firing, $m$ and $I_{zz}$ denote the system's mass and inertia. Since two respective thrusters are actuated together, a firing creates $2F_\text{n}$ force.
Note that the input matrix $\mathbf{B}(\theta)$ depends on the orientation of the system, so it is a linear time-variant system.
\subsection{Systems limit cycle}
The minimum activation time for a thruster causes the system to experience a minimum thrust force or acceleration.
Consequently, this leads to a minimum change in velocity on the system:
\begin{equation}
    \Delta v_{\text{min}} = \frac{2F_\text{n} t_\text{on,min}}{m}\label{eq:delta-v-min}
\end{equation}
During this minimum firing, the system will travel a certain distance:
\begin{equation}
       \Delta x_\text{min} = \displaystyle + \frac{2F_\text{n} \left(t_\text{on,min}\right)^{2}}{2m} + t_\text{off,min} v_{0} + t_\text{on,min} v_{0} \label{eq:delta-x-bound} \\
\end{equation}
Where $v_0$ denotes the system's velocity before the firing, measured on the thruster axis and the direction of thrust force. Note the assumption that the thruster had just fired before, and hence waits $t_\text{off,min}$.
If the system has a speed lower than $\Delta v_{\text{min}}$ on the axis in which respective thruster points, a minimum thrust firing overcompensates. The system will start moving in the other direction, with a velocity that is again smaller than $ \Delta v_{\text{min}}$. Hence, the system can not be fully stopped \citep{mendel_performance_1968}.

Instead of fully stopping the system, it can only be held in an artificial limit cycle around the target position. Assuming that an optimal controller fires at the right moment to keep the system as close as possible, an upper one-dimensional position bound for the limit cycle can be derived:
\begin{equation}
    \pm \left( \frac{2F_\text{n} t_{\text{off, min}} t_\text{on,min}}{m} + \frac{0.625\cdot2F_\text{n} \left(t_\text{on,min}\right)^{2}}{m}\right)
\end{equation}
\section{MPC formulation}\label{sec:mpc}
The general \acrshort{mpc} optimization problem is stated in the following:
\begin{subequations}
\begin{gather}
   \min_{\textbf{x}_{t+k|t}, \textbf{x}_{t+N|t}, \textbf{u}_{t+k|t}} \mathcal{L}_f(\textbf{x}_{t+N|t})+\sum^{N-1}_{k=0}\mathcal{L}(\textbf{x}_{t+k|t}, \textbf{u}_{t+k|t})\label{eq:mpc-cost}
\end{gather}
\begin{align}
    \forall k \in [0,N)~\text{s.t.}~ & \textbf{x}_{t+k+1|t} = f(\textbf{x}_{t+k|t}, \textbf{u}_{t+k|t}), \label{eq:mpc-sysdyn}\\
    & \textbf{u}_{t+k|t} \in \mathbb{U}  \label{eq:input_cons}\\
    & \textbf{x}_{t+k|t} \in \mathbb{X}  \label{eq:mpc-cons-state}\\
    & \textbf{x}_{t+N|t} \in \mathbb{X}_f\label{eq:mpc-final-cons-state}\\
    & \textbf{x}_{t|t} = \textbf{x}_t
\end{align}
\end{subequations}
Note, that $\textbf{x}_{t+k|t}$ and $\textbf{u}_{t+k|t}$ refers to the $k$ steps into the future predicted state or input at time step $t$.
The above problem consists of the cost function \eqref{eq:mpc-cost} to minimize. The system prediction is based on the system dynamics constraint~\eqref{eq:mpc-sysdyn}. The inputs $\textbf{u}_{t+k|t}$ have to be in the input set $\mathbb{U}$~\eqref{eq:input_cons}, the intermediate predicted states within the state set $\mathbb{X}$~\eqref{eq:mpc-cons-state} and the final predicted state within the final state set $\mathbb{X}_f$~\eqref{eq:mpc-final-cons-state}.
In the following, the parts of the optimization problem are explained in more detail.
The three \acrshort{mpc} formulations compared in this work are listed within Table~\ref{tab:forms} and consist of different parts.
\begin{table}[htpb]
    \small
    \centering
    \ra{1.3}
    \begin{tabular}{@{}llll@{}}
    \toprule
         & \textbf{Binary constraints} & \textbf{Cost function}\\
         \hline
     \acrshort{milp} & Integer constraint & L1 Norm\\
     \acrshort{qp} & Penalty term & L2 Norm\\
     \acrshort{mpcc} & \acrshort{lcc} & L2 Norm\\
     \bottomrule
    \end{tabular}
    \caption{The three different \acrshort{mpc} foundations that are compared in this work together with their optimization problem ingredients.}
    \label{tab:forms}
    \vspace{-0.4cm}
\end{table}

\subsection{System dynamics}
The system dynamics~\eqref{eq:sys_dyn} are linearized using the first-order Taylor expansion around the current state~$\mathbf{x_t}$. This is equivalent to evaluating the state-dependent input Matrix $\mathbf{B}(\theta)$ at the current orientation $\theta_t$. The system is discretized using the backward Euler approach, with a sampling rate of $\Delta t$. Hence, the general dynamic constraint~\eqref{eq:mpc-sysdyn} in the \acrshort{mpc} formulation expands to a set of linear constraints:
\setcounter{MaxMatrixCols}{20}
\begin{align}
    \mathbf{x}_{t+k+1|t} = \mathbf{x}_{t+k|t} +& \Delta t \mathbf{A} \mathbf{x}_{t+k+1|t}\nonumber + \Delta t \mathbf{B} \mathbf{u}_{t+k|t}
\end{align}
Where $\mathbf{A}$ and $\mathbf{B} = \mathbf{B}(\theta_t)$ refer to the state and input matrix of the system dynamics~\eqref{eq:sys_dyn}.
\subsection{Thruster timing constraints}
\begin{figure}[htpb]
    \centering
    \includegraphics[width=0.65\linewidth]{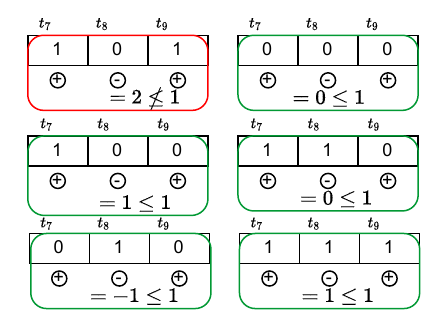}
    \vspace{-0.5cm}
    \caption{Example for the input sequence (top left) that violates the minimum off-time constraint and sequences that don't. }
    \label{fig:min_off}
\end{figure}

\paragraph{Minimum on time}
The chosen system discretization rate $\Delta t = 0.1\si{\second}$ matches the minimum on time  $t_{\text{on, min}} = 0.1\si{\second}$. Hence, by assuming a zero-order hold, the minimum on-time is enforced naturally.
\paragraph{Maximum on time}
The maximum on time is a multiple of the discretization rate: $t_{\text{on, max}} = 3 \cdot \Delta t = 0.3\si{\second}$. 
Therefore, a constraint must prevent the binary input variables from having the value $1$ for four consecutive time steps, respectively. This is done via a sliding window constraint, which for each input limits the sum of all consecutive subsequences of length four to three and is stated as:
\begin{multline}
    \sum_{j = k}^{k + 3} u_{i,t+j|t} \leq 3,
    \forall k \in [-3,N - 3), \forall i \in [1,3] \label{eq:cons_max_on}
\end{multline}
\begin{figure}[htpb]
    \centering
    \includegraphics[width=0.65\linewidth]{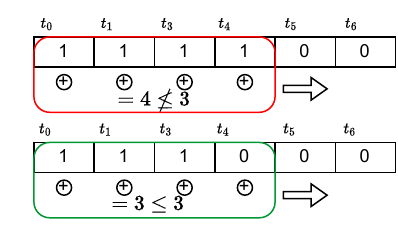}
    \vspace{-0.25cm}
    \caption{Example for an input sequence that (top) violates the maximum on time constraint and (bottom) does not.}
    \label{fig:max_on}
\end{figure}
The sliding summation window constraint is sketched for a sequence of 6 binary inputs in Fig.~\ref{fig:max_on}.
\paragraph{Minimum off time}
The minimum off time is twice the discretization rate, i.e. $t_{\text{off, min}} = 2 \cdot \Delta t = 0.2\si{\second}$.
Hence, an input value of $1$ must be followed by another $1$ or two consecutive $0$. Hence, the input sequence $(1,0,1)$ is prevented by another sliding window constraint (Fig.~\ref{fig:min_off}), which is given by:
\begin{gather}
    u_{i,t+k-1|t} - u_{i,t+k|t} + u_{i,t+k+1|t}\leq 1,~\forall k \in [-2,N-1), \nonumber \\ \forall i \in [1,3]\label{eq:cons_min_off}
\end{gather}
\subsection{Continuous input constraints}
The applied torque is limited by a maximum torque:
\begin{align}
    -\tau_\text{max} \leq \textbf{u}_{0,t+k|t} \leq \tau_\text{max} \label{eq:cons_cont_in}
\end{align}
\subsection{Binary input constraints}
The three formulations to enforce binary values for the thruster inputs compared in this work are stated in the following.
Together with the bounding box constraint on the continuous input~\eqref{eq:cons_cont_in}, they replace the input constraint 
\eqref{eq:input_cons} in the \acrshort{mpc} formulation.
\paragraph{\acrlong{lcc}}
The condition for the input variables only taking value $0$ or $1$ can be expressed as a set of \acrshort{lcc}'s:
\begin{align}
    &  0 \leq (1 - u_{i,t+k|t}) \perp u_{i,t+k|t} \geq 0, ~\forall i \in [1,3]\label{eq:lc}
\end{align}
These constraints emerge from \acrlong{lcp}s and the "$\perp$" means that either the left or the right term must be zero, while both have to be non-negative \citep{cottle_linear_2009}. A \acrlong{mpcc} is difficult \todo[inline]{cite and give NP-hardness?} to solve, and there exist many techniques and specialized solvers for \acrshort{mpcc} specializations \citep{hall_sequential_2022}. Note that \eqref{eq:lc} can be reformulated into a general \acrfull{nlp}:
\begin{subequations}
\label{eq:lc_nlp}
\begin{gather}
    (1 - u_{i,t+k|t}) \enspace u_{i,t+k|t} = 0\\
    0 \leq u_{i,t+k|t} \leq  1\label{eq:lc_lin2}
\end{gather}
\end{subequations}
This formulation can be solved with any appropriate \acrshort{nlp} solver \citep{fletcher_solving_2004}.
With the \acrshort{lcc}, the optimization problem becomes a \acrfull{lcqp}. 

\paragraph{Penalty term}
In this formulation, the binary variables are not enforced via a hard constraint. Instead, a quadratic penalizing term~\cite{abudia_switched_2020} together with a bounding box constraint is added to the cost function:
\begin{subequations}
\begin{align}
    J^*(x_t) =& \min_{U_t, X_t} \mathcal{L}_f(\textbf{x}_{t+N|t})+\sum^{N-1}_{k=0}\mathcal{L}(\textbf{x}_{t+k|t}, \textbf{u}_{t+k|t}) \nonumber\\
    +  \sum^3_{j=1} 4 \beta \enspace & (u_{i,t+k|t} -  u_{i,t+k|t}^2),~  \beta > 0 \label{eq:big-m-cost}\\
        & 0 \leq u_{i,t+i|t} \leq 1, \forall i \in [1,3]\label{eq:big-m-bb}
\end{align}
\end{subequations}
The penalty term is only optimal if the respective variable is $0$ or $1$.
Note that the penalizing term is non-convex. Hence, the optimization problem becomes a non-convex \acrshort{qp} and can be solved by any appropriate \acrshort{qp}-solver.

\paragraph{Integer constraint}
The binary decision variables are formulated as integer variables which can only take the value $0$ or $1$:
\begin{align}
    &  u_{i,t+k|t} \in \{0,1\}, ~\forall i \in [1,3]\label{eq:int-cons}
\end{align}
This approach transforms the problem into a \acrlong{mip}. As mentioned above, solving these problems requires a special class of solver.
\subsection{State constraints}
The system state is limited by a lower- and upper bound to stay below a safety velocity, and not to exceed the edges of the flat floor. Thus~\eqref{eq:mpc-cons-state} becomes:
\begin{equation}
    \mathbf{x}_\text{lb} \leq \textbf{x}_{t+k|t} \leq \mathbf{x}_\text{ub}\\
\end{equation}
\subsection{Final constraint}
To ensure recursive feasibility, the terminal constraint~\eqref{eq:mpc-final-cons-state} is set to the velocity limits of the minimum limit cycle~\eqref{eq:delta-v-min}.
Hence, it can be ensured that after the final prediction step, the system can be kept within the limit cycle bounds~\eqref{eq:delta-x-bound}.

\subsection{Cost function}
The cost function \eqref{eq:mpc-cost} is used in two ways. First, to minimize the deviation from the target pose $\mathbf{\Hat{x}}\in\mathbbm{R}^6$. Secondly, to minimize the control effort, in this case mainly thrust.
In this work, due to the different binary constraint formulations and solvers used, two different cost terms are studied:
\paragraph{L1 Norm}
The L1 norm can be expressed as a linear sum of additional linear auxiliary constraints:
\begin{subequations}
\begin{align}
 \mathcal{L}(\mathbf{x}_{t+k|t} - \mathbf{\Hat{x}}, \mathbf{u}_{t+k |t}) &=  \mathbbm{1}^T\mathbf{e}_{x,k} + \mathbbm{1}^T\mathbf{e}_{u,k}\\
 \mathcal{L}_f(\mathbf{x}_{t+N|t} - \mathbf{\Hat{x}}) &=  \mathbbm{1^T}\mathbf{e}_{x,N} \label{eq:l1_cost}
\end{align}
 \end{subequations}
 Where $\mathbf{e}_{x,t}\in\mathbbm{R}^6, \mathbf{e}_{u,t}\in\mathbbm{R}^5$ are the auxiliary vectors for state and input cost. They are related to the state and input by additional linear constraints:
 \begin{subequations}
 \begin{gather}
  -\mathbf{e}_{x,k} \leq \mathbf{Q}(\mathbf{x}_{t+k|t} - \mathbf{\Hat{x}}) \leq + \mathbf{e}_{x,k} \label{eq:l1_cons}\\
   -\mathbf{e}_{u,k} \leq \mathbf{W}(\mathbf{u}_{t+k|t}) \leq + \mathbf{e}_{u,k}
\end{gather}
 \end{subequations}
Where $\mathbf{Q}\in\mathbbm{R}^{6\times},  \mathbf{W}\in\mathbbm{R}^{5\times},$ are the cost matrices that define the weighting for each state and input. The auxiliary vectors are added as another decision variable to the optimization problem.
Hence, the optimization problem stays a linear program.
\paragraph{L2 Norm}
The L2 norm is expressed as a quadratic cost:
\begin{align}
 \mathcal{L}(\mathbf{x}_{t+k|t} - \mathbf{\Hat{x}}, \mathbf{u}_{t+k |t}) &= (\mathbf{x}_{t+k|t} - \mathbf{\Hat{x}})^T\mathbf{Q}(\mathbf{x}_{t+k|t} - \mathbf{\Hat{x}}) \nonumber \\
 &+ (\mathbf{u}_{t+k|t})^T\mathbf{W}(\mathbf{u}_{t+k|t})\\
 \mathcal{L}_f(\mathbf{x}_{t+N|t} - \mathbf{\Hat{x}}) &=  (\mathbf{x}_{t+N|t} - \mathbf{\Hat{x}})^T\mathbf{Q}(\mathbf{x}_{t+N|t} - \mathbf{\Hat{x}}) \label{eq:l2_cost}
\end{align}
\section{Results}\label{sec:results}
The controllers have been tested on a rigid body simulation using \textit{drake toolbox} \citep{tedrake_drake_2019}. \textit{SNOPT solver} \citep{gill_snopt_2005} is used to solve the non-convex \acrshort{qp} and \acrshort{mpcc}. In addition, the solver \textit{LCQPow} \citep{hall_sequential_2022} represents a specialized \acrshort{lcqp} solver. To efficiently solve the \acrshort{mip} problem, \textit{SCIPSolver} \citep{bestuzheva_scip_2021}  with python bindings \citep{maher_pyscipopt_2016} is used. All experiments were performed on standard hardware with a \nth{12} generation I7 processor and \SI{32}{\giga\byte} of RAM.

\subsection{Feasible region}
The feasible regions of the different formulations are calculated to compare their ability to find solutions. 
The initial state is $\textbf{x}_0 = [x_0, \SI{0}{m},\SI{45}{\degree},\dot x_0,\SI{0}{\frac{\meter}{\second}}, \SI{0}{\frac{\radian}{\second}}]^T$, where $x_0, \dot x_0$ take different values to analyze the feasibility in state space.
The final velocity constraint is set to the limit cycle bounds~\eqref{eq:delta-v-min}. Position error and thruster usage are minimized via the cost function.
\begin{figure}[htpb]
    \centering
    \includegraphics{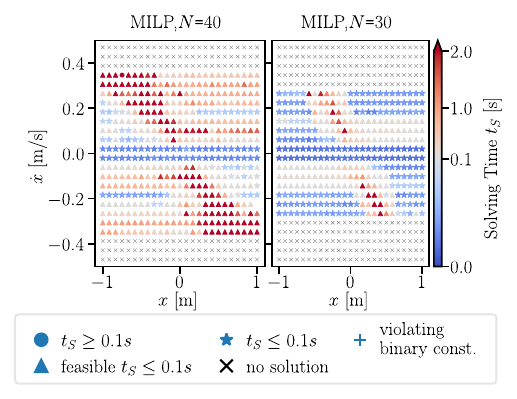}
    \vspace{-0.6cm}
    \caption{Feasible region for \acrshort{milp} under final limit cycle velocity constraint. Position error and thruster usage are minimized via cost.}
    \label{fig:mi_cost}
\end{figure}
Fig.~\ref{fig:mi_cost} shows the feasible region for the \acrshort{milp} formulation for different prediction horizons.
For each initial condition, the feasible solutions are marked according to their solving time $t_s$. A star indicates $t_s \leq 0.1s$, which was chosen as the desired real-time solving rate. Due to the final velocity constraint, the maximum initial $\dot x$, for which the solver finds a solution, is limited. For large values, the solver is not able to enter the velocity limit cycle within the prediction horizon. A higher prediction horizon leads to a bigger feasible region.
For a prediction horizon $N=30$, the solving time is up to \SI{2}{\second} for certain regions, indicated by the red areas in Fig.~\ref{fig:mi_cost}. For the longer prediction horizon, this becomes even more evident. However, the triangles indicate that there was at least one feasible suboptimal solution available at $t_s\leq0.1s$.
\begin{figure}[htpb]
    \centering
    \includegraphics{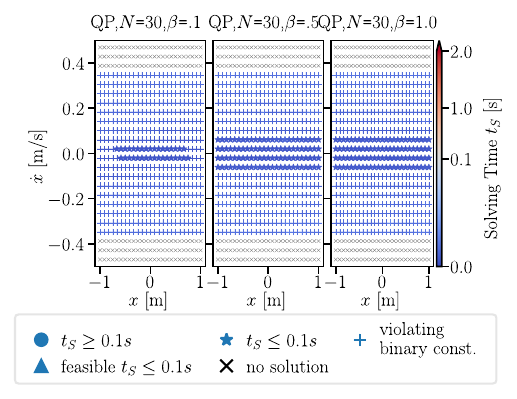}
    \vspace{-0.7cm}
    \caption{Feasible region for \acrshort{qp} under final limit cycle velocity constraint. Position error and thruster usage are minimized via cost. The different plots refer to different weightings of the penalty cost.}
    \label{fig:bigm_cost}
\end{figure}\\
Fig.~\ref{fig:bigm_cost} shows the feasible regions for the \acrshort{qp} formulation for a prediction horizon of $N=40$ and different weightings of the penalty term.
The feasible regions are bigger and the solving time is significantly lower than for the \acrshort{milp}. However, the plus indicates that for most of the initial conditions, the penalty costs couldn't be minimized. Hence, the input values do not take a binary value. No significant improvement can be observed through higher weighting of the penalty term.

Experiments with the \acrshort{mpcc} have shown that for a few initial conditions, a solution can be found mostly below $0.05s$. However, most of the time, all solvers fail to find a solution. One option is to relax the constraint. However, choosing the relax parameter, in general, is not practical because it introduces non-binary values to the solution.
\todo[inline]{Add plots of optimality difference?}
\todo[inline]{Add phase plots?}

Since the objective of this paper is an \acrshort{mpc} that optimally controls the system while respecting binary and timing constraints, only the \acrshort{milp} formulation is appropriate. 
The \acrshort{lcp} formulation provides only partial or no solution, while the \acrshort{qp} formulation often yields non-binary inputs, which may require rounding up or down in a preliminary step. The timing constraints and theoretical assumptions however assume binary inputs, which means that in reality, these constraints will not be met.
Therefore, the \acrshort{milp} formulation is chosen. If an optimal solution is not found within \SI{0.1}{\second}, suboptimal solutions are used. The impact of suboptimally (i.e., prediction horizon) on suboptimally is analyzed in the following.
\subsection{Closed loop simulation}
\begin{figure}[htpb]
    \centering
    \includegraphics{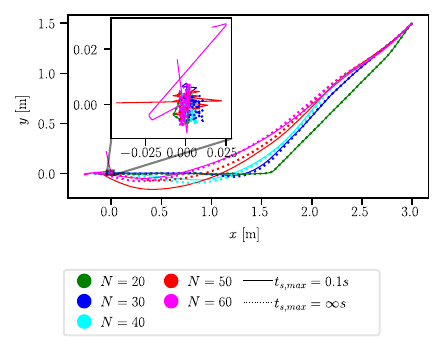}
    \vspace{-0.5cm}
    \caption{System trajectories under \acrshort{mpc} control law for different prediction horizons for optimal solver solutions and suboptimal, fast solutions.}
    \label{fig:opti_N_cmp}
\end{figure}
\begin{figure}[htpb]
    \centering
    \includegraphics{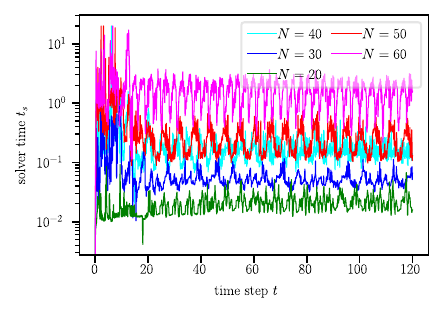}
    \vspace{-0.5cm}
    \caption{Solving time for \acrshort{milp} solver to find the optimal solution over the whole simulation for different prediction horizons.}
    \label{fig:opti_N_cmp_time}
\end{figure}

Fig.~\ref{fig:opti_N_cmp} shows the trajectory of the system for the closed-loop simulation for different prediction horizons. For each horizon, there exist two trajectories. One where the optimal result is taken as the control input and one where the most optimal solution within \SI{0.1}{\second} is taken, to meet the real-time requirement.
The initial system state is $x=[3.0m, 1.5m, 90^\circ, 0,0,0]^T$, and the target to minimize it towards the origin (i.e., $x=[0, 0, 0, 0,0,0]^T$).
Fig.~\ref{fig:opti_N_cmp_time} shows the time that the solver took to find the optimal solution for each time step for the different prediction horizons.
Only horizon $N=20$ never exceeds the target limit of \SI{0.1}{\second}. For $N=30$ the optimal solution was found not within \SI{0.1}{\second} for some time steps at the beginning. For the rest of the experiment, the optimal solution could be found within \SI{0.1}{\second}. 
For $N=40$ the solver time oscillates around \SI{0.1}{\second}, while $N=50$ requires especially at the beginning more considerably more time. For $N=60$, the solver takes on average more than \SI{1}{\second} to find the optimal solution. Note that for all these tests, a feasible solution was always found within \SI{0.1}{\second}.
The different prediction horizons also reflect in the system trajectories obtained by the respective controllers (Fig.\ref{fig:opti_N_cmp}). It is worth noting that for values of $N$ up to 40, the suboptimal and optimal trajectories are similar. On the other hand, for $N=50$, a distinct overshoot becomes evident. For $N=60$ the system position at the origin can not be maintained, and it drifts away.

\begin{figure}[htpb]
    \centering
    \includegraphics{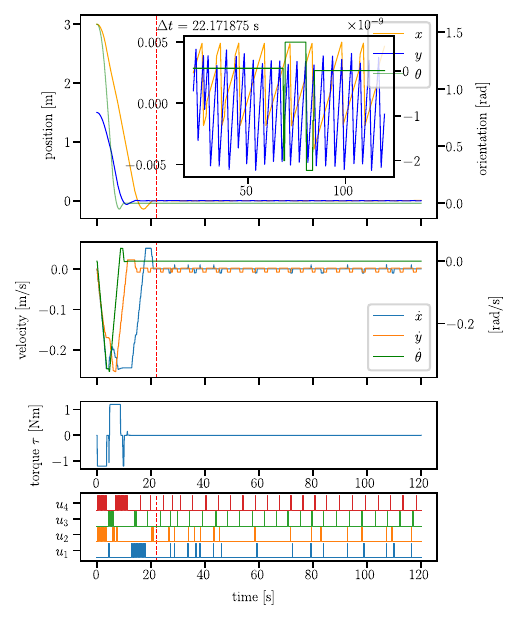}
    \vspace{-1cm}
    \caption{States and input of the system controlled by the \acrshort{milp} \acrshort{mpc} towards the origin, where position is maintained.}
    \label{fig:syst_traj-N40}
\end{figure}

A more detailed plot of the system's behavior together with actuation for $N=40$ under \acrshort{milp} controller is shown in Fig.~\ref{fig:syst_traj-N40}. It shows that the system is steered within \SI{22}{\second} to the origin, where it is oscillating in a limit cycle with \SI{\pm 0.004}{\meter} around the origin.
\todo[inline]{Maybe relate to calculations from above.}
The thruster usage at the beginning is high, to accelerate the system and finally break it into the limit cycle. Where then a much lower thruster usage is necessary to maintain the limit cycle. 
\section{Summary and Conclusion}\label{sec:summary_conc}
The comparison of the three different binary constraint formulations shows that only the \acrshort{milp} formulation works in practice. 
Even if the non-convex \acrshort{qp} finds optimal results, most of the time the binary constraints are violated. A higher weighting of the penalty function can not counteract this. 
Either way, this formulation adds another tuning parameter to the problem, which has a critical impact. In addition, the constraints are not guaranteed to be met.\\
The \acrshort{mpcc} formulations were very promising, since the constraints are enforced, whereas the problem can be solved in theory with any non-linear solver. 
However, it shows that the \acrshort{lcp} constraints in this problem are too complex to solve for the standard solver. Also, a special solver tested in this work could not bring any improvement. It is worth noting that there are other solvers available for comparison in future work. However, these solvers often use \acrshort{mip} solving techniques and the results are likely similar.\\
Although the \acrshort{milp} formulation may not always yield the optimal solution in a reasonable time, it has been demonstrated that it can always identify at least one feasible solution. Therefore, it can be assumed that the controller system always stays feasible. Additionally, it was found that the feasible solution, which is obtained within \SI{0.1}{\second}, has a narrow optimality gap for a sufficiently small prediction horizon.
The simulation experiments demonstrate that the \acrshort{milp} \acrshort{mpc} formulation can effectively guide the system to the origin and maintain it there.\\
Furthermore, this work shows a simple linear formulation of the binary input timing constraints. For future work, it would be interesting if these constraints can be generalized and applied to other timings.\\
The next steps are to extend this to a coupled formulation involving all thrusters and the reaction wheel limits, to fully exploit the capabilities of the system. Secondly, the \acrshort{mpc} has to be tested on the real system to determine if it can handle model errors and disturbances by the not perfectly even flat floor.

\section*{Acknowledgments}
The first and second authors would like to acknowledge the support of M-RoCk (Grant No.: FKZ 01IW21002) and AAPLE (Grant Number: 50WK2275) and the second author would also like to acknowledge the support from the European Union’s Horizon 2020 research and innovation programme under the Marie Skłodowska-Curie grant agreement No 813644. 
The first author would like to thank his university, Universität zu Lübeck, and especially the Institute for Electrical Engineering in Medicine, for their support.
Furthermore, many thanks to the \textit{drake}, \textit{SCIPSolver} and \textit{PySCIPOpt} development teams for their tools and active support. 
\small
\bibliography{references}

\end{document}

%% file: acronyms.tex
\usepackage[acronym]{glossaries}

\makeglossaries

\newacronym{orgl}{ORGL}{Orbital Robotics and GNC Lab}
\newacronym{esa}{ESA}{European Space Agency}
\newacronym{asdr}{ASDR}{Active Space Debris Removal}
\newacronym{reacsa}{REACSA}{REcap-ACrobat-SAtsim}
\newacronym{tvlqr}{TVLQR}{Time-Varying Linear Quadratic Regulator}
\newacronym{mpc}{MPC}{Model Predictive Control} 
\newacronym{pwm}{PWM}{Pulse-width modulation} 
\newacronym{lp}{LP}{Linear Program} 
\newacronym{nlp}{NLP}{Non Linear Program} 
\newacronym{qp}{QP}{Quadratic Program} 
\newacronym{mip}{MIP}{Mixed Integer Program} 
\newacronym{milp}{MILP}{Mixed Integer Linear Program} 
\newacronym{mi}{MI}{Mixed Integer} 
\newacronym{miqp}{MIQP}{Mixed Integer Quadratic Program} 
\newacronym{scip}{SCIP}{Good question what does this mean} 
\newacronym{lcp}{LCP}{Linear Complementarity Program} 
\newacronym{lcqp}{LCQP}{Linear Complementarity Quadratic Program} 
\newacronym{lpcc}{LPCC}{Linear Program with Complementarity Constraints} 
\newacronym{mpcc}{MPCC}{Mathematical Program with Complementarity Constraints} 
\newacronym{lcc}{LCC}{Linear Complementarity Constraints} 
\newacronym{rw}{RW}{Reaction Wheel}
\newacronym{mimpc}{MIMPC}{Mixed Integer Model Predictive Control}

%% file: main.bbl
\begin{thebibliography}{21}
\providecommand{\natexlab}[1]{#1}
\providecommand{\url}[1]{\texttt{#1}}
\expandafter\ifx\csname urlstyle\endcsname\relax
  \providecommand{\doi}[1]{doi: #1}\else
  \providecommand{\doi}{doi: \begingroup \urlstyle{rm}\Url}\fi

\bibitem[Abudia et~al.(2020)Abudia, Harlan, Self, and
  Kamalapurkar]{abudia_switched_2020}
Moad Abudia, Michael Harlan, Ryan Self, and Rushikesh Kamalapurkar.
\newblock Switched {Optimal} {Control} and {Dwell} {Time} {Constraints}: {A}
  {Preliminary} {Study}.
\newblock In \emph{2020 59th {IEEE} {Conference} on {Decision} and {Control}
  ({CDC})}, pages 3261--3266, Jeju, Korea (South), December 2020. IEEE.
\newblock ISBN 978-1-72817-447-1.
\newblock \doi{10.1109/CDC42340.2020.9304087}.
\newblock URL \url{https://ieeexplore.ieee.org/document/9304087/}.

\bibitem[Arantes et~al.(2009)Arantes, Martins-Filho, and
  Santana]{arantes_optimal_2009}
Gilberto Arantes, Luiz~S. Martins-Filho, and Adrielle~C. Santana.
\newblock Optimal {On}-{Off} {Attitude} {Control} for the {Brazilian}
  {Multimission} {Platform} {Satellite}.
\newblock \emph{Mathematical Problems in Engineering}, 2009:\penalty0 e750945,
  November 2009.
\newblock ISSN 1024-123X.
\newblock \doi{10.1155/2009/750945}.
\newblock URL \url{https://www.hindawi.com/journals/mpe/2009/750945/}.

\bibitem[Bestuzheva et~al.(2021)Bestuzheva, Besançon, Chen, Chmiela,
  Donkiewicz, Doornmalen, Eifler, Gaul, Gamrath, Gleixner, Gottwald, Graczyk,
  Halbig, Hoen, Hojny, Hulst, Koch, Lübbecke, Maher, Matter, Mühmer, Müller,
  Pfetsch, Rehfeldt, Schlein, Schlösser, Serrano, Shinano, Sofranac, Turner,
  Vigerske, Wegscheider, Wellner, Weninger, and Witzig]{bestuzheva_scip_2021}
Ksenia Bestuzheva, Mathieu Besançon, Wei-Kun Chen, Antonia Chmiela, Tim
  Donkiewicz, Jasper~van Doornmalen, Leon Eifler, Oliver Gaul, Gerald Gamrath,
  Ambros Gleixner, Leona Gottwald, Christoph Graczyk, Katrin Halbig, Alexander
  Hoen, Christopher Hojny, Rolf van~der Hulst, Thorsten Koch, Marco Lübbecke,
  Stephen~J. Maher, Frederic Matter, Erik Mühmer, Benjamin Müller, Marc~E.
  Pfetsch, Daniel Rehfeldt, Steffan Schlein, Franziska Schlösser, Felipe
  Serrano, Yuji Shinano, Boro Sofranac, Mark Turner, Stefan Vigerske, Fabian
  Wegscheider, Philipp Wellner, Dieter Weninger, and Jakob Witzig.
\newblock The {SCIP} {Optimization} {Suite} 8.0.
\newblock {ZIB}-{Report} 21-41, Zuse Institute Berlin, December 2021.
\newblock URL \url{http://nbn-resolving.de/urn:nbn:de:0297-zib-85309}.

\bibitem[Bredenbeck et~al.(2022)Bredenbeck, Vyas, Zwick, Borrmann,
  Olivares-Mendez, and Nüchter]{bredenbeck_trajectory_2022}
Anton Bredenbeck, Shubham Vyas, Martin Zwick, Dorit Borrmann, Miguel
  Olivares-Mendez, and Andreas Nüchter.
\newblock Trajectory {Optimization} and {Following} for a {Three} {Degrees} of
  {Freedom} {Overactuated} {Floating} {Platform}.
\newblock \emph{2022 IEEE/RSJ International Conference on Intelligent Robots
  and Systems (IROS)}, pages 4084--4091, July 2022.
\newblock URL \url{http://arxiv.org/abs/2207.10693}.
\newblock arXiv:2207.10693 [cs].

\bibitem[Cottle et~al.(2009)Cottle, Pang, and Stone]{cottle_linear_2009}
Richard~W. Cottle, Jong-Shi Pang, and Richard~E. Stone.
\newblock \emph{The {Linear} {Complementarity} {Problem}}.
\newblock Classics in {Applied} {Mathematics}. SIAM, 2 edition, 2009.
\newblock ISBN 978-0-89871-686-3.
\newblock \doi{10.1137/1.9780898719000}.

\bibitem[Doman et~al.(2007)Doman, Gamble, and Ngo]{doman_control_2007}
David Doman, Brian Gamble, and Anhtuan Ngo.
\newblock Control {Allocation} of {Reaction} {Control} {Jets} and {Aerodynamic}
  {Surfaces} for {Entry} {Vehicles}.
\newblock In \emph{{AIAA} {Guidance}, {Navigation} and {Control} {Conference}
  and {Exhibit}}, Guidance, {Navigation}, and {Control} and {Co}-located
  {Conferences}. American Institute of Aeronautics and Astronautics, August
  2007.
\newblock \doi{10.2514/6.2007-6778}.
\newblock URL \url{https://arc.aiaa.org/doi/10.2514/6.2007-6778}.

\bibitem[Fletcher and Leyffer(2004)]{fletcher_solving_2004}
Roger Fletcher and Sven Leyffer.
\newblock Solving mathematical programs with complementarity constraints as
  nonlinear programs.
\newblock \emph{Optimization Methods and Software}, 19\penalty0 (1):\penalty0
  15--40, February 2004.
\newblock ISSN 1055-6788.
\newblock \doi{10.1080/10556780410001654241}.
\newblock URL \url{https://doi.org/10.1080/10556780410001654241}.

\bibitem[Gill et~al.(2005)Gill, Murray, and Saunders]{gill_snopt_2005}
Philip~E. Gill, Walter Murray, and Michael~A. Saunders.
\newblock {SNOPT}: {An} {SQP} {Algorithm} for {Large}-{Scale} {Constrained}
  {Optimization}.
\newblock \emph{SIAM Review}, 47\penalty0 (1):\penalty0 99--131, 2005.
\newblock \doi{10.1137/S0036144504446096}.
\newblock URL \url{https://doi.org/10.1137/S0036144504446096}.

\bibitem[Hall et~al.(2022)Hall, Nurkanović, Messerer, and
  Diehl]{hall_sequential_2022}
Jonas Hall, Armin Nurkanović, Florian Messerer, and Moritz Diehl.
\newblock A {Sequential} {Convex} {Programming} {Approach} to {Solving}
  {Quadratic} {Programs} and {Optimal} {Control} {Problems} {With} {Linear}
  {Complementarity} {Constraints}.
\newblock \emph{IEEE Control Systems Letters}, 6:\penalty0 536--541, 2022.
\newblock ISSN 2475-1456.
\newblock \doi{10.1109/LCSYS.2021.3083467}.

\bibitem[Khayour et~al.(2020)Khayour, Durand, Cuvillon, and
  Gangloff]{khayour_active_2020}
Imane Khayour, Sylvain Durand, Loïc Cuvillon, and Jacques Gangloff.
\newblock Active {Damping} of {Parallel} {Robots} {Driven} by {Elastic}
  {Cables} using {On}-{Off} {Actuators} through {Model} {Predictive} {Control}
  {Allocation}.
\newblock \emph{IFAC-PapersOnLine}, 53\penalty0 (2):\penalty0 9169--9174,
  January 2020.
\newblock ISSN 2405-8963.
\newblock \doi{10.1016/j.ifacol.2020.12.2167}.
\newblock URL
  \url{https://www.sciencedirect.com/science/article/pii/S2405896320328202}.

\bibitem[Leeuwen(1990)]{leeuwen_algorithms_1990}
Jan Leeuwen.
\newblock \emph{Algorithms and {Complexity}}, volume~1 of \emph{Handbook of
  {Theoretical} {Computer} {Science}}.
\newblock Elsevier, September 1990.
\newblock ISBN 978-0-444-88071-0.

\bibitem[Leomanni et~al.(2013)Leomanni, Garulli, Giannitrapani, and
  Scortecci]{leomanni_mpc-based_2013}
Mirko Leomanni, Andrea Garulli, Antonio Giannitrapani, and Fabrizio Scortecci.
\newblock An {MPC}-based attitude control system for all-electric spacecraft
  with on/off actuators.
\newblock In \emph{52nd {IEEE} {Conference} on {Decision} and {Control}}, pages
  4853--4858, December 2013.
\newblock \doi{10.1109/CDC.2013.6760650}.
\newblock ISSN: 0191-2216.

\bibitem[Maher et~al.(2016)Maher, Miltenberger, Pedroso, Rehfeldt, Schwarz, and
  Serrano]{maher_pyscipopt_2016}
Stephen~J. Maher, Matthias Miltenberger, João~Pedro Pedroso, Daniel Rehfeldt,
  Robert Schwarz, and Felipe Serrano.
\newblock {PySCIPOpt}: {Mathematical} {Programming} in {Python} with the {SCIP}
  {Optimization} {Suite}.
\newblock volume 9725, pages 301--307. Springer, 2016.
\newblock \doi{10.1007/978-3-319-42432-3_37}.

\bibitem[Mendel(1968)]{mendel_performance_1968}
Jerry Mendel.
\newblock Performance cost functions for a reaction-jet-controlled system
  during an on-off limit cycle.
\newblock \emph{IEEE Transactions on Automatic Control}, 13\penalty0
  (4):\penalty0 362--368, August 1968.
\newblock ISSN 1558-2523.
\newblock \doi{10.1109/TAC.1968.1098940}.
\newblock Conference Name: IEEE Transactions on Automatic Control.

\bibitem[Posa et~al.(2014)Posa, Cantu, and Tedrake]{posa_direct_2014}
Michael Posa, Cecilia Cantu, and Russ Tedrake.
\newblock A direct method for trajectory optimization of rigid bodies through
  contact.
\newblock \emph{The International Journal of Robotics Research}, 33\penalty0
  (1):\penalty0 69--81, January 2014.
\newblock ISSN 0278-3649.
\newblock \doi{10.1177/0278364913506757}.
\newblock URL \url{https://doi.org/10.1177/0278364913506757}.

\bibitem[Sopasakis et~al.(2015)Sopasakis, Bernardini, Strauch, Bennani, and
  Bemporad]{sopasakis_hybrid_2015}
Pantelis Sopasakis, Daniele Bernardini, Hans Strauch, Samir Bennani, and
  Alberto Bemporad.
\newblock A {Hybrid} {Model} {Predictive} {Control} {Approach} to {Attitude}
  {Control} with {Minimum}-{Impulse}-{Bit} {Thrusters}.
\newblock In \emph{2015 {European} {Control} {Conference} ({ECC})}, pages
  2079--2084, July 2015.
\newblock \doi{10.1109/ECC.2015.7330846}.

\bibitem[Tedrake and Drake-Development-Team(2019)]{tedrake_drake_2019}
Russ Tedrake and Drake-Development-Team.
\newblock Drake: {Model}-based design and verification for robotics, 2019.
\newblock URL \url{https://drake.mit.edu}.

\bibitem[Vieira et~al.(2011)Vieira, Galvão, and Kienitz]{vieira_attitude_2011}
Márcio~Santos Vieira, Roberto Kawakami~Harrop Galvão, and Karl~Heinz Kienitz.
\newblock Attitude stabilization with actuators subject to switching-time
  constraints using explicit {MPC}.
\newblock In \emph{2011 {Aerospace} {Conference}}, pages 1--8, March 2011.
\newblock \doi{10.1109/AERO.2011.5747482}.

\bibitem[Virgili-Llop et~al.(2018)Virgili-Llop, Zagaris, Park, Zappulla, and
  Romano]{virgili-llop_experimental_2018}
Josep Virgili-Llop, Costantinos Zagaris, Hyeongjun Park, Richard Zappulla, and
  Marcello Romano.
\newblock Experimental evaluation of model predictive control and inverse
  dynamics control for spacecraft proximity and docking maneuvers.
\newblock \emph{CEAS Space Journal}, 10\penalty0 (1):\penalty0 37--49, March
  2018.
\newblock ISSN 1868-2510.
\newblock \doi{10.1007/s12567-017-0155-7}.
\newblock URL \url{https://doi.org/10.1007/s12567-017-0155-7}.

\bibitem[Vyas et~al.(2022)Vyas, Maywald, Kumar, Jankovic, Mueller, and
  Kirchner]{vyas_post-capture_2022}
Shubham Vyas, Lasse Maywald, Shivesh Kumar, Marko Jankovic, Andreas Mueller,
  and Frank Kirchner.
\newblock Post-capture detumble trajectory stabilization for robotic active
  debris removal.
\newblock \emph{Advances in Space Research}, September 2022.
\newblock ISSN 0273-1177.
\newblock \doi{10.1016/j.asr.2022.09.033}.
\newblock URL
  \url{https://www.sciencedirect.com/science/article/pii/S0273117722008742}.

\bibitem[Zwick et~al.(2018)Zwick, Huertas, Gerdes, and Ortega]{zwick_orgl_2018}
Martin Zwick, Irene Huertas, Levin Gerdes, and Guillermo Ortega.
\newblock {ORGL} – {ESA}’s {Test} {Facility} for {Approach} and {Contact}
  operations in {Orbital} and {Planetary} {Environments}.
\newblock In \emph{Proceedings of the {International} {Symposium} on
  {Artificial} {Intelligence}, {Robotics} and {Automation} in {Space}
  (i-{SAIRAS})}, volume~6, Madrid, Spain, June 2018.

\end{thebibliography}
